\documentclass[runningheads]{llncs}
\usepackage[T1]{fontenc}
\usepackage{array}
\usepackage{algorithm}
\usepackage{algorithmicx}
\usepackage{algpseudocode}
\usepackage{amsmath}
\usepackage{amssymb}
\usepackage{orcidlink}
\usepackage[utf8]{inputenc} 
\usepackage{pifont} 

\newcommand{\cmark}{\ding{51}} 
\newcommand{\xmark}{\ding{55}} %

\usepackage{graphicx}
\usepackage{xcolor}
\usepackage{tikz}

\usepackage{hyperref}
\usepackage{color}

\begin{document}
%
\title{Hallucination Detection in Large Language Models using Diversion Decoding}
\titlerunning{Diversion Decoding for Hallucination Detection}
%
\author{Basel Abdeen\inst{1}\orcidlink{0009-0002-4059-1928} \and
S M Tahmid Siddiqui\inst{1}\orcidlink{0009-0007-1931-0569} \and
Meah Tahmeed Ahmed\inst{1}\orcidlink{0009-0003-0972-1998} \and
Anoop Singhal\inst{2}\orcidlink{0000-0002-2602-3927} \and
Latifur Khan\inst{1}\orcidlink{0000-0002-9300-1576}
\and
Punya Parag Modi\inst{1} \orcidlink{0009-0008-1640-2596}
\and
Ehab Al-Shaer\inst{3}\orcidlink{0000-0002-7665-8293
}
}
\authorrunning{B. Abdeen et al.}
%
\institute{The University of Texas at Dallas, Richardson, TX 75080, USA
\email{\{basel.abdeen,tahmid.siddiqui,meah.ahmed,lkhan\}@utdallas.edu} 
\and
National Institute of Standards and Technology, Gaithersburg, USA
\email{anoop.singhal@nist.gov}
\and
Carnegie Mellon University, PA, USA
\email{ehab@cmu.edu}
}
%
\maketitle              
\begin{tikzpicture}[remember picture,overlay]
\node[anchor=north west,align=center,text width=\textwidth,font=\footnotesize]
  at ([xshift=\dimexpr\oddsidemargin+1in\relax,yshift=-0.5cm]current page.north west)
  {Accepted for publication in DBSec 2025. The final publication is 
  
  available at Springer via \url{https://doi.org/10.1007/978-3-031-96590-6_7}.};
\end{tikzpicture}
\begin{abstract}
Large language models (LLMs) have emerged as a powerful tool for retrieving knowledge through seamless, human-like interactions. Despite their advanced text generation capabilities, LLMs exhibit hallucination tendencies, where they generate factually incorrect statements and fabricate knowledge, undermining their reliability and trustworthiness. Multiple studies have explored methods to evaluate LLM uncertainty and detect hallucinations. However, existing approaches are often probabilistic and computationally expensive, limiting their practical applicability.

In this paper, we introduce diversion decoding, a novel method for developing an LLM uncertainty heuristic by actively challenging model-generated responses during the decoding phase. Through diversion decoding, we extract features that capture the LLM's resistance to produce alternative answers and utilize these features to train a machine-learning model to develop a heuristic measure of the LLM's uncertainty. Our experimental results demonstrate that diversion decoding outperforms existing methods with significantly lower computational complexity, making it an efficient and robust solution for evaluating hallucination detection.

\end{abstract}
\section{Introduction}
Recently, large language models (LLMs) have gained the world's attention as they started to exhibit a deep understanding of natural language and a solid grasp of the world's knowledge, prompting people to use them for knowledge retrieval instead of traditional search engines~\cite{brown2020language,10.5555/3600270.3602446,long2024generative}. Although LLMs have demonstrated high performance in various tasks, their reliability as a source of information is limited due to their tendency to hallucinate \cite{huang2025survey,ji2023survey}. Hallucination is LLM behavior that arises when a model is uncertain of its knowledge, leading it to invent facts and generate fake information. 

Multiple studies have attempted to tackle the LLM hallucination challenge by introducing various scores to develop heuristic measures for LLM uncertainty~\cite{farquhar2024old,fu-etal-2024-gptscore,ji2023survey,Farquhar2024}. These studies proposed approaches that require an external knowledge base, use probabilistic approaches, or exhibit high computational complexity. Unlike these studies, this paper presents a deterministic and computationally efficient approach for quantifying LLMs' confidence without requiring any external components beyond an existing question-answering dataset. Our approach is inspired by state-of-the-art methodologies that leverage the relationship between consistency and confidence exhibited by LLMs~\cite{farquhar2024old,Farquhar2024,manakul-etal-2023-selfcheckgpt}. When an LLM is confident, it tends to generate the correct answer in various syntactic forms while maintaining semantic similarity. Conversely, when the LLM lacks confidence, it produces semantically different answers for the same question through different decoding paths.

Unlike previous studies where multiple answers need to be sampled from an LLM~\cite{farquhar2024old,Farquhar2024,manakul-etal-2023-selfcheckgpt}, our approach, diversion decoding, requires generating two distinct answers as follows. We first prompt the LLM with a question and retrieve the greedy answer. Then, we prompt the LLM with the same question; however, whenever the LLM generates an answer that is semantically similar to the first one, we steer its generation toward a different answer. For a simple example, we ask the LLM, "What is the capital of France?" Assume that the first greedy answer is "Paris." We then prompt the LLM with the same question, but this time, whenever the LLM generates the word "Paris," we reject this generation and select the next possible token with the highest likelihood. We hypothesize that when the LLM is confident, it will keep trying to generate the same answer; otherwise, it is easy to steer it away toward a different answer. We captured various features of LLM behavior during diversion decoding and trained a machine learning model to develop a proxy measure of LLM uncertainty based on a labeled dataset. In our experiments, diversion decoding demonstrated robust performance in detecting hallucinations while requiring substantially less computational power compared to existing state-of-the-art approaches.

The remainder of this paper is structured as follows: Section 2 presents the motivation behind our approach. Section 3 defines the problem statement. Section 4 introduces the proposed diversion decoding method. Section 5 discusses the evaluation results. Section 6 reviews related work, and finally, Section 7 provides the conclusion, including limitations and directions for future research.

\section{Motivation}

Large language models (LLMs) have gained significant traction among individuals and organizations for various applications, including translation, summarization, chatbots, and knowledge retrieval. These applications range in complexity from simple question-answering systems to sophisticated customer assistance agents. Despite the widespread applicability of LLMs, there remain significant concerns about their reliability and trustworthiness \cite{kang2023deficiency}. One of the primary challenges to their reliability stems from their hallucination tendencies, where LLMs generate fabricated or nonfactual information.
Table~\ref{ex_hall} illustrates an example of LLM hallucination in an open question-answering context.

In many use cases, incorrect decision-making by LLM-powered agents can result in significant financial loss or business disruption \cite{kang2023deficiency}. For example, LLMs are increasingly being integrated into automated cybersecurity systems for tasks such as malware detection, security automation, and cyber forensics \cite{ferrag2025generative}. Hallucinations in such systems can lead to inaccurate threat assessments, false alarms, or security misconfigurations that expose vulnerabilities. An LLM mistakenly flagging benign activity as malicious may cause disruptions, while failing to detect a genuine attack could leave critical systems exposed. Moreover, in regulated environments, where data accuracy and confidentiality are mandated by laws such as the General Data Protection Regulation (GDPR) and the Health Insurance Portability and Accountability Act (HIPAA), LLM hallucinations can result in compliance violations, legal repercussions, and reputational damage \cite{das2025security}.

Despite ongoing efforts by AI experts to raise awareness about the risks of LLM hallucinations, users often accept LLM-generated outputs at face value, making them susceptible to fake information \cite{azaria-mitchell-2023-internal,huang2025survey}. As a result, detecting and mitigating hallucinations remains a critical challenge in integrating LLMs into high-stakes applications. Effective methods for measuring and identifying hallucinations are essential for ensuring the reliability of LLMs in domains such as cybersecurity, healthcare, and finance. The stronger these detection mechanisms become, the more confidently LLMs can be deployed in critical decision-making systems.

\begin{table}[t]
    \centering
    \renewcommand{\arraystretch}{1.3} 
    \caption{Example of LLM Hallucination from a Bard promotion tweet posted by Google in 2023 \cite{Coulter_Bensinger_2023} }
    \begin{tabular}{| m{3cm} | m{8cm} |}
        \hline
        \textbf{Prompt} & What new discoveries from the James Webb Space Telescope (JWST) can I tell my 9-year-old about? \\ 
        \hline
        \textbf{LLM Response} & "...JWST took the very first pictures of a planet outside of our own solar system: the first-ever exoplanet image..." \\ 
        \hline
        \textbf{Correct Info} & The first pictures of exoplanets were taken by the European Southern Observatory's Very Large Telescope (VLT) in 2004, as confirmed by NASA. \\ 
        \hline
    \end{tabular}
    \label{ex_hall}
\end{table}

\section{Problem Statement}
In this paper, we address the challenge of detecting LLM hallucinations by quantifying their confidence in their responses to factual questions. We introduce diversion decoding, a novel approach that leads to a proxy measure of an LLM's confidence in its greedy answer by deliberately attempting to steer it toward an alternative semantically different response. We use the degree of resistance exhibited by the model in deviating from its initial answer as a proxy measure of uncertainty. To quantify this resistance, we train a supervised machine learning model on a public dataset of factual question-answer pairs to predict whether a generated answer is true or hallucinated using features derived from diversion decoding. By quantifying LLM resistance, our approach provides a reliable estimation for LLM hallucination. Similar to recent studies~\cite{farquhar2024old,Farquhar2024}, this work aims to use a proxy measure for the semantic uncertainty of an entire LLM-generated answer, beyond the uncertainty of individual tokens.

Our approach focuses on open question-answering tasks for factual questions. Unlike free-response questions, factual questions require a single, concise answer. The approach is particularly suited for developing proxy measures of uncertainty in open-source LLMs, where output token probabilities are accessible. Finally, we employ supervised learning to train a machine learning model on an existing question-answering dataset. This dataset plays a crucial role in distinguishing instances when LLM generate correct answers from those when they hallucinate by analyzing patterns in LLM behavior.


\section{Approach}

Diversion decoding helps develop a proxy for the LLM's uncertainty in answering a question by first generating the greedy answer and then forcing the model to produce a different response by diverting it away from the greedy answer during decoding. This process tests how "adamant" the LLM is about its initial response and, consequently, how certain it is of that answer.

\begin{figure*}[t]

\centering
\includegraphics[width=10cm]{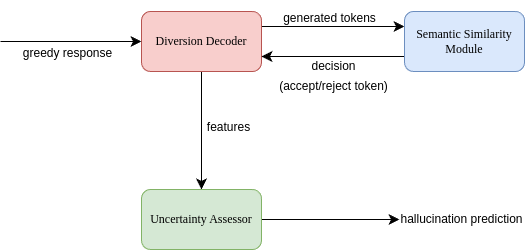}
\caption{Diversion decoding pipeline}
\label{fig:module_flowchart}
\end{figure*}

Diversion decoding comprises three key components: a diversion decoder, a semantic similarity module, and an uncertainty assessment module. The diversion decoder prompts the LLM with a question, retrieves the greedy answer, and then compels the model to generate a second, distinct response. The semantic similarity module detects when the LLM attempts to reproduce the greedy answer and alerts the decoder to steer the generation away from that response. Finally, the uncertainty assessment module assesses the LLM's uncertainty based on the path it takes to generate the second answer. Figure~\ref{fig:module_flowchart} illustrates a high-level overview of the diversion decoding pipeline.

\begin{figure*}[t]

\centering
\includegraphics[width=12cm]{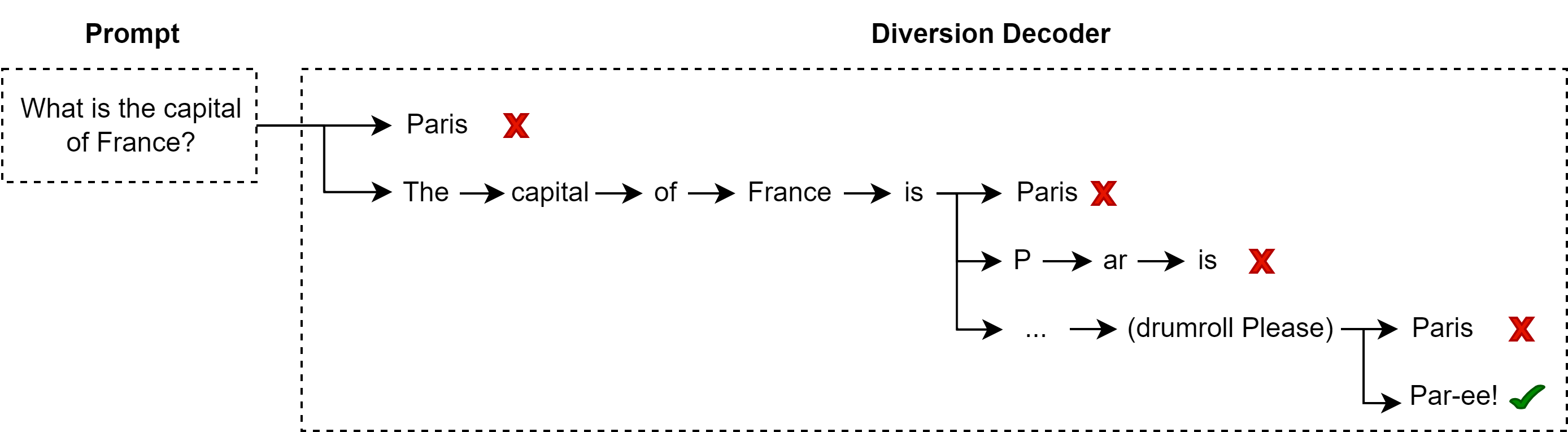}
\caption{Diversion decoding example when prompting the LLM with the question "What is the capital of France?" and getting Paris as the greedy answer. The LLM keeps trying to generate the answer "Paris" through different generation paths, demonstrating
its confidence in the answer.}
\label{fig:example1}
\end{figure*}

\begin{figure*}[t]

\centering
\includegraphics[width=12cm]{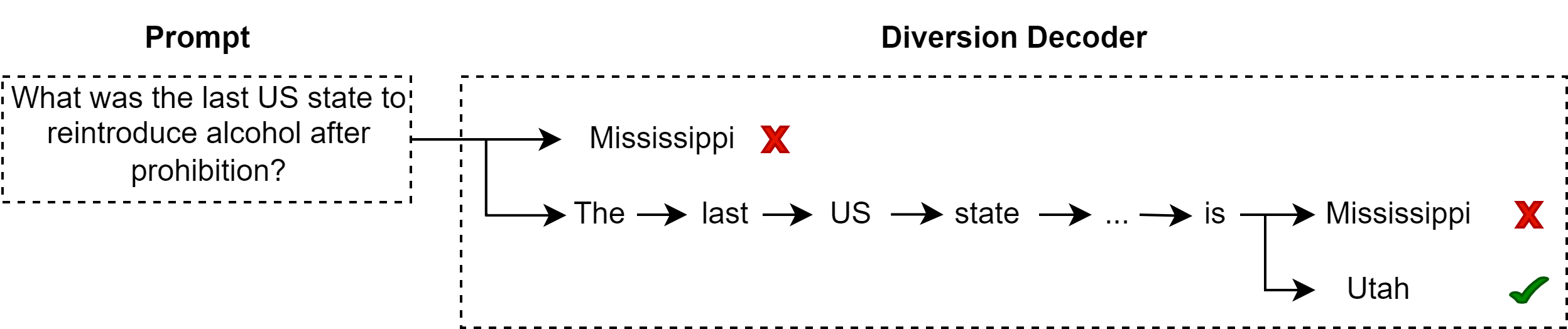}
\caption{Diversion decoding example when prompting the LLM with the question "What is the last US state to reintroduce alcohol after prohibition?" and getting Mississippi as the greedy answer. The LLM modified its answer from Mississippi to Utah after two rejections from the similarity module, demonstrating low confidence in its answer.}
\label{fig:example2}
\end{figure*}

Figure~\ref{fig:example1} illustrates diversion decoding in action for the question, "What is the capital of France?" Each time the decoder generates "Paris," the similarity module rejects the response, forcing the decoder to explore alternative generation paths. Since this is a straightforward question that an LLM should be highly confident in answering, the model persistently attempts to generate "Paris" through different paths. In contrast, Figure~\ref{fig:example2} demonstrates a case where the LLM was easily diverted from its greedy answer (Mississippi) to an alternative response (Utah). Unlike in the first example, where the LLM repeatedly attempted to generate the same answer through different routes, here, it readily produced a completely different response.

\begin{algorithm}
\caption{Diversion Decoder Algorithm}
\begin{algorithmic}[1]
\State \textbf{Input:} Large Language Model LLM, Question $Q$, Top-k parameter $k$, Similarity Threshold $\delta$

\State \textbf{Output:} Greedy Answer $G$ and a Distinct Answer $R$
\State Prompt LLM with $Q$ and generate a greedy answer $G$
\State Re-prompt LLM with $Q$ 
\State Initialize $R \gets \emptyset$
\While{End-of-response token not reached}
    \State Instruct LLM to generate probability distribution for next token
    \State Select top-$k$ tokens with highest probabilities
    \For{each token $t$ in top-$k$}
        \State Append the greedy word of $t$ to $R$ 
        \State Get $R$ similarity to $G$
        \If{Similarity < $\delta$}
            \State Append the greedy word of $t$ to $R$
            \State \textbf{break}
        \EndIf
    \EndFor
    \If{No token in top-$k$ passes similarity check}
        \State Terminate response generation
    \EndIf
\EndWhile
\State \textbf{Return} $G$ and $R$
\end{algorithmic}
\end{algorithm}

\subsection{Diversion decoding}

\noindent{\bf LLM decoding background} LLMs generate a response to a prompt one token at a time. Given an input text, it generates an output token, appends it to the input, and then generates the next token. This process continues until the model produces a special token indicating the end of the response. At each step, the LLM generates a probability distribution over its predefined vocabulary. Based on a selected decoding strategy, the model chooses a token from this distribution. In greedy decoding, the LLM selects the token with the highest probability at each step. In top-k decoding, the model randomly selects a token from the k most probable tokens. In top-p (nucleus) decoding, the LLM samples a token from the smallest set of tokens whose cumulative probability mass reaches p. Moreover, sampling temperature modifies the Softmax distribution over possible next tokens. A higher temperature flattens the distribution, increasing randomness by making lower-probability tokens more likely. A lower temperature sharpens the distribution, concentrating probability on the most likely tokens, leading to more predictable and conservative outputs. In diversion decoding, we always select the greedy answer, which correlates to choosing top-k equals one or temperature equals zero. 

\noindent{\bf Diversion decoding} The diversion decoder first prompts the LLM with a question and generates a greedy answer. It then re-prompts the LLM with the same question but enforces the generation of a different response using the following approach. The decoder instructs the LLM to generate the probability distribution for the next token and selects the top-k tokens with the highest probabilities. It then attempts to generate a response by selecting the most probable token. If the similarity component approves the selected token, the decoder appends it to the prompt and proceeds to generate the next token. This process continues until the end-of-response token is reached. Conversely, if the similarity component rejects the generated token, the decoder selects the next highest probability token from the top-k candidates. This iterative process continues until either the similarity component approves a token, or all top-k tokens are exhausted, with none passing the similarity check. More formally, the token \( \tilde{r}_i \) at decoding step \( i \) is selected according to the following rule:

\begin{equation}
\tilde{r}_i = \arg\max_{\substack{t \in \text{TopK}_i \\ \text{sim}(t, G) < \delta}} P_i\left(t \mid Q, \tilde{r}_1, \ldots, \tilde{r}_{i-1}\right)
\label{eq:diversion_decode}
\end{equation}

Where:
\begin{itemize}
    \item \( Q \): The input prompt.
    \item \( G \): The greedy answer.

    \item \( P_i(t \mid \cdot) \): The model's predicted probability of token \( t \) at step \( i \).
    \item \( \text{TopK}_i \): The top-\( k \) highest-probability tokens at step \( i \)
    \item \( \text{sim}(t, G) \): The similarity between generated answer R with token \( t \)  and the greedy answer G.
    \item \( \delta \): The similarity threshold.
\end{itemize}

Algorithm [1] showcases how our diversion decoding algorithm works. In lines 3-4, we prompt an LLM with a question, $Q$, to generate a greedy answer, $G$, and then we re-prompt the model with $Q$. We then begin an iterative process in line 6, which continues until the end-of-response token is reached. For each iteration, we generate the LLM's probability distribution for the next token (line 7), and we select the top $k$ tokens with the highest probabilities (line 8). Here, $k$ is a predefined value that is provided to the algorithm. We then enter an inner iterative loop that goes over each token in the queue of top $k$ tokens. For each token, we form a greedy word by appending to it the highest likelihood tokens till the first space (line 10). This word-forming approach ensures that the semantic similarity module has the complete word context for accurate semantic comparison. We then extract the similarity score between the greedy answer $G$ and the distinct answer $R$ (line 11) and compare it to a predefined threshold, $\delta$ (line 12). If the similarity score is below the defined threshold, we append the token with its greedy word to our distinct answer, $R$ (line 13), and break from the inner loop. Finally, if no token from the top-k queue passes the similarity check, we terminate the entire diversion decoding algorithm. The algorithm outputs the greedy answer, $G$, and the distinct answer, $R$.

\noindent{\bf Similarity Measure}
The similarity component is designed to detect whether the decoder generates an answer that is semantically similar to the greedy answer. We implement a comprehensive similarity function that incorporates three distinct similarity metrics: sentence semantic similarity, word semantic similarity, and character-based similarity. These metrics were chosen based on their complementary strengths in capturing different aspects of textual similarity.
Let $A$ represent the generated answer concatenated with the next token candidate, and $G$ represent the first greedy answer. The number of tokens in the greedy answer is denoted by $L$.
We define auxiliary functions $last(t,n)$ and $first(t,n)$ to extract the last $n$ words and first $n$ words from text $t$, respectively. The function $encode(t)$ represents the semantic encoding of text $t$ using all-mpnet-base-v2 model. The function $cos(x,y)$ computes the cosine similarity between vectors $x$ and $y$.
The complete similarity function $sim(A,G)$ is composed of several components. The first component, shown in Equation (2), computes the sentence-level semantic similarity:
\begin{equation}
semantic\_sim1(A,G) = cos(encode(last(A,L+1)), encode(G))
\end{equation}
This semantic similarity metric captures the overall semantic relationship between the generated and greedy answers.
The second component, defined in Equation (3), focuses on word-level comparisons:
\begin{equation}
semantic\_sim2(A,G) = cos(encode(last(A,1)), encode(first(G,1)))
\end{equation}

The character-based similarity component, shown in Equation (4), employs the longest common subsequence (LCS) algorithm:
\begin{equation}
character\_sim(A,G) = \frac{LCS(A,G)}{Length(G)}
\end{equation}
These three components are combined in the final similarity score computation, as shown in Equation (5):
\begin{equation}
\begin{aligned}
    sim(A,G) = \max(&\text{semantic\_sim1}(A,G), \\
                    &\text{semantic\_sim2}(A,G), \\
                    &\text{character\_sim}(A,G))
\end{aligned}
\end{equation}

Equation (2) addresses the common phenomenon where LLMs generate introductory or framing text before providing the actual answer. To handle this behavior, we consider only the last L+1 words of the generated response. This is particularly evident in diversion decoding, where token rejection often leads the LLM to generate additional, semantically irrelevant introductory text. As demonstrated in Figure~\ref{fig:example1}, LLMs may produce verbose outputs such as "The capital of France is...(drumroll please)... Par-ee." By restricting our semantic comparison as shown in Equation (2), we achieve significantly higher accuracy compared to analyzing the entire response.

Equation (3) serves a complementary purpose by focusing specifically on the first non-stop word of the greedy answer and the last generated token of the current generation. This approach proves particularly effective when dealing with lengthy greedy answers, as it enables early intervention in the decoding process before a substantial portion of the answer is generated. 

The character-based similarity metric defined in Equation (4) addresses an observed behavior in LLMs where, when forced to alter their response, they may generate concatenated words without proper spacing. This phenomenon can confound purely semantic similarity models, leading to false negatives in similarity detection. For instance, when the expected answer is "North Carolina," the LLM might generate "NorthCarolina" - a response that maintains character-level similarity despite lacking proper formatting. The LCS metric in Equation (4) successfully captures these cases.

The maximum-based aggregation in Equation (5) ensures that the highest similarity detected by any of the metrics is preserved in the final score, making the system more robust to various types of semantic equivalence. This multi-metric approach to similarity measurement demonstrates superior performance compared to single-metric approaches.

\subsection{Uncertainty Assessment}
After the decoder generates the second answer, the uncertainty assessment component aims to assign a persistence score to the greedy answer based on how resilient the LLM was against generating a different answer. Multiple attributes of the diversion decoding process can capture the LLM's uncertainty about its answer. For example, more rejections indicate that the model is more certain of its answer. However, in many cases, other features are essential to consider. For example, not all rejected tokens should be treated equally, as a rejected token with a higher likelihood demonstrates LLM certainty better than a rejected token with a low likelihood. Moreover, the LLM can have fewer rejections because it generates the end-of-response token before generating a full answer due to its certainty of the greedy answer.

Instead of using one metric to assess hallucination, we train a machine learning model using a dataset of open question-answer pairs - the TriviaQA dataset - as follows. First, we prompted the LLM with a question and obtained its greedy answer. The greedy answer is labeled correct if its ROUGE-L~\cite{lin2004rouge}
similarity with the ground truth exceeded 0.3. Otherwise, it is considered a hallucination. 

Second, we use diversion decoding to generate an answer that is semantically different from the greedy answer. Third, we collect the following six features from the diversion decoder: (1) the number of rejected tokens, (2) the sum of the negative log-likelihood of the rejected tokens, (3) the
minimum negative log-likelihood of a rejected token, (4) the maximum negative
log-likelihood of approved tokens, (5) the mean negative log-likelihood of all
generated tokens, and (6) the length of the largest common substring between
the greedy answer and the second-best answer. Finally, we applied this process to all questions in the dataset, and we trained a gradient-boosting model to predict if the greedy answer is hallucinated based on the extracted features and labels. 



\begin{table}[t]

    \centering
    \renewcommand{\arraystretch}{1.2}
    \caption{Example question-answer pair from our modified dataset file}
    \begin{tabular}{| m{3cm} | m{8cm} |}
        \hline
        \textbf{Question} & Where in England was actor Nigel Hawthorne born? \\ 
        \hline
        \textbf{Aliases} & "Cofantre", "Coventry (city)", "Coventry, Warwickshire", "Coventry", "Coventry, UK", "Coventry, England", "City of Coventry", "COVENTRY", "County Borough of Coventry", "Coventry (borough)", "Coventry City Council", "Coventry, United Kingdom", "Metropolitan Borough of Coventry" \\ 
        \hline
        \textbf{Matched Wiki Entity Name} & Coventry \\ 
        \hline
        \textbf{Normalized Aliases} & aliases converted to lowercase and stripped of special characters and punctuation \\ 
        \hline
        \textbf{Normalized Value} & coventry \\ 
        \hline
        \textbf{Type} & WikipediaEntity \\ 
        \hline
        \textbf{Value} & Coventry \\ 
        \hline
    \end{tabular}
    \label{ex_question}
\end{table}

\section{Evaluation}
\subsection{Dataset, model and metrics}

We evaluate diversion decoding using TriviaQA, a large-scale open question-answering dataset~\cite{joshi-etal-2017-triviaqa}. TriviaQA was authored by trivia enthusiasts and contains a wide range of trivia questions in various domains, styles, and subjects (e.g., history, science, and culture), ensuring a comprehensive evaluation of diversion decoding.

The TriviaQA dataset is organized into two primary folders: one for reading comprehension and another for question answering. The unfiltered-web-dev.json file from the question-answering folder (version 1.0) contains 11,313 question-answer pairs, which were curated by trivia enthusiasts and supported by independently gathered evidence documents. We utilized 4,900 questions from this particular file. Table~\ref{ex_question} shows a sample from the dataset. Each question-answer pair consists of the question, the answer (value), variations of the correct answer (aliases), normalized versions of the answers, and metadata about the source of the answer (matched wiki entity name and the type of source). 

We selected the Llama family of models~\cite{touvron2023Llama} for our experiments because of its open-source accessibility, diversity in model sizes, and comparable performance to state-of-the-art closed-source models. Moreover, we used a 4-bit quantization of Llama to reduce memory usage and computational cost. Our experiments were all done using a single NVIDIA H100 GPU. We leveraged the Area Under the Receiver Operating Characteristic Curve (AUROC) for performance evaluation. Rather than defining a specific hallucination threshold, AUROC assesses the model’s ability to differentiate between true positive and false positive classifications across a range of decision thresholds.

\subsection{Experiments}
\noindent\textbf{Baselines} 
We evaluated diversion decoding against four baselines: predictive entropy \cite{kadavath2022language}, length-normalized predictive entropy \cite{malinin2020uncertainty}, lexical similarity \cite{fomicheva2020unsupervised}, and semantic entropy~\cite{farquhar2024old,Farquhar2024}. The predictive entropy is the conditional entropy of the generated tokens given the context. 
\begin{equation} H(W \mid X) = - \sum_{w \in V} P(w \mid X) \log P(w \mid X) \end{equation} where \( X \) represents the given context, \( W \) is the next token to be generated, and \( V \) is the model’s vocabulary. In sequential generation, the entropy at each decoding step can be summed across the sequence to get the joint entropy: \begin{equation} H(Y \mid X) = \sum_{t=1}^{T} H(W_t \mid X, W_{<t}) \end{equation}

Length-Normalized Predictive Entropy adjusts the predictive entropy of longer sequences by dividing the joint log probability of each generated sequence by its length. Normalizing asserts that the expected uncertainty of generations is independent of sentence length. 
Lexical similarity measures the average similarity between the answers in a generated answer set by computing the average Rouge-L score for all pairs of sentences. Semantic entropy focuses on the meaning of text rather than the specific tokens used. It addresses "semantic equivalence," where different sentences can have identical meanings. The process involves sampling sequences from a language model, clustering sequences with the same meaning using a bi-directional entailment algorithm, and finally, calculating entropy over the distribution of meanings, summing probabilities of semantically equivalent sequences.

\noindent\textbf{Training} We randomly selected 1,300 questions from the TriviaQA dataset to train our uncertainty estimation component. We trained a gradient-boosting classifier as we discussed in the diversion decoding section. We used log loss, a learning rate of 0.1, 100 estimators and the mean squared error with improvement score by Friedman to measure the quality of a split.

\begin{table}[t]
\caption{Diversion decoding and baselines results}
\renewcommand{\arraystretch}{1.2}
\begin{tabular}{>{\centering\arraybackslash\footnotesize}m{18mm}|
>{\centering\arraybackslash\footnotesize}m{18mm}|
>{\centering\arraybackslash\footnotesize}m{18mm}|
>{\centering\arraybackslash\footnotesize}m{18mm}|
>{\centering\arraybackslash\footnotesize}m{18mm}|
>{\centering\arraybackslash\footnotesize}m{18mm}}
\hline 

Model & Predictive entropy & Normalized predictive entropy & Lexical similarity & Semantic entropy & Diversion decoding \\ \hline
Llama 7B & 67.0\% & 67.8\% & 64.4\% & 71.9\% & \textbf{74.66\%} \\ \hline
Llama 13B & 70.1\% & 63.8\% & 61.7\% & 72.1\% & \textbf{78.49\%}
\\ \hline 

\end{tabular}
\label{results}

\end{table}

\begin{table}[t]
\caption{Diversion decoding and baselines expansion ratio assuming n=10 for lexical similarity and semantic entropy}
\begin{tabular}{>{\centering\arraybackslash\footnotesize}m{18mm}|
>{\centering\arraybackslash\footnotesize}m{18mm}|
>{\centering\arraybackslash\footnotesize}m{18mm}|
>{\centering\arraybackslash\footnotesize}m{18mm}|
>{\centering\arraybackslash\footnotesize}m{18mm}|
>{\centering\arraybackslash\footnotesize}m{18mm}}
\hline 

 & Predictive entropy & Normalized predictive entropy & Lexical similarity & Semantic entropy & Diversion decoding \\ \hline
Expansion ratio & 1 & 1 & 10 & 10 & 3.6 \\ \hline

\end{tabular}
\label{comp}

\end{table}

\noindent\textbf{Results}
We selected 3,600 random questions from TriviaQA to evaluate diversion decoding against the other baselines. We used Llama 2 models of 7 and 13 billion parameters. Table~\ref{results} presents the experimental results across different evaluation metrics, comparing the performance of Llama 7B and Llama 13B models. For the Llama 7B model, diversion decoding outperforms all other approaches, achieving an AUROC score of 74.66\%. The second-best approach is semantic entropy, which attains a 71.9\% score, followed closely by predictive entropy at 67.0\%. Normalized predictive entropy and lexical similarity yield slightly lower performance at 67.8\% and 64.4\%, respectively.
Similarly, for the Llama 13B model, diversion decoding again demonstrates superior performance, reaching an AUROC of 78.49\%, showing a notable improvement over the 7B variant. Semantic entropy remains competitive with a score of 72.1\%, maintaining consistency across model sizes. Predictive entropy and lexical similarity yield scores of 70.1\% and 61.7\%, respectively, while normalized predictive entropy shows a slightly lower performance at 63.8\%.

\noindent\textbf{Complexity analysis}
To assess the complexity of our approach compared to other methods, we compute the ratio of all generated tokens to the number of tokens in the initial response. We refer to this metric as the expansion ratio (ER), as it quantifies how much the generated response expands relative to the initial response. 
 More formally, let \( T_{\text{init}} \in \mathbb{N} \) denote the number of tokens in the initial response, and let \( T_{\text{gen}} \in \mathbb{N} \) denote the total number of tokens after generation. The \emph{Expansion Ratio} (ER) is defined as:

\begin{equation}
\mathrm{ER} := \frac{T_{\text{gen}}}{T_{\text{init}}}
\end{equation}
While both our approach and existing methods use auxiliary models for semantic similarity or natural language inference, these models contain significantly fewer parameters than the large language model and can be safely ignored in our calculations.

In diversion decoding, the number of generated tokens varies across different questions based on the confidence of the large language model, as discussed in the diversion decoding section. To account for this variation, we empirically compute the ratio of generated tokens to the number of tokens in the greedy answer and report the median across our dataset of 4,900 examples. 
The results are presented in Table~\ref{comp}. Predictive entropy and normalized predictive entropy do not require any extra generation beyond the first response, thus, their expansion ratio is 1.
On average, diversion decoding achieves an expansion ratio of 3.6, which is significantly lower than approaches that sample n responses, where the expansion ratio is n. For example, when 10 samples are sampled for lexical
similarity or semantic entropy, the expansion ratio is 10.

\begin{figure}[t]

\centering
\includegraphics[width=10cm]{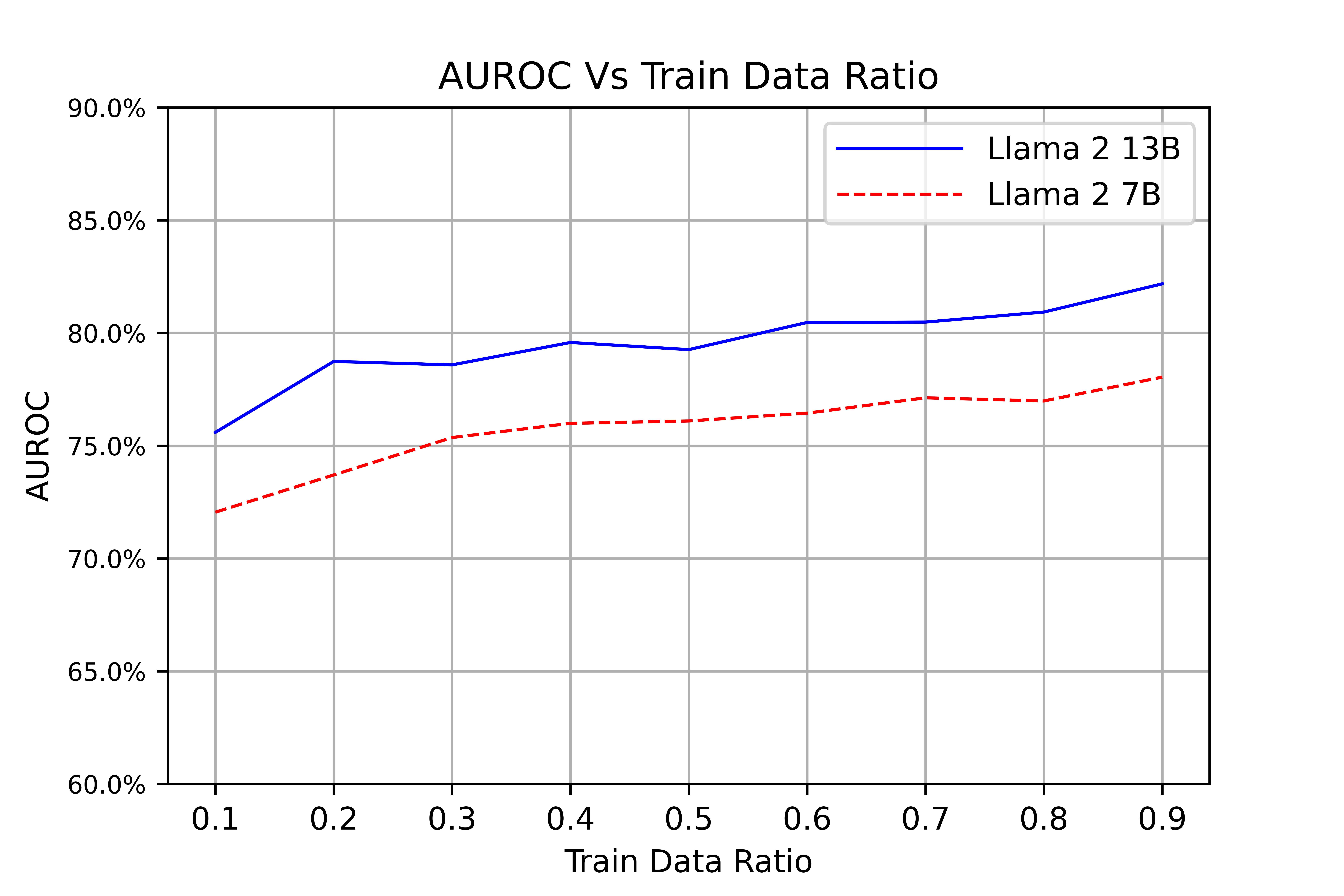}
\caption{Diversion decoding 
AUROC vs training data ratio}
\label{fig:train_ratio}
\end{figure}

\begin{figure}[t]

\centering
\includegraphics[width=10cm]{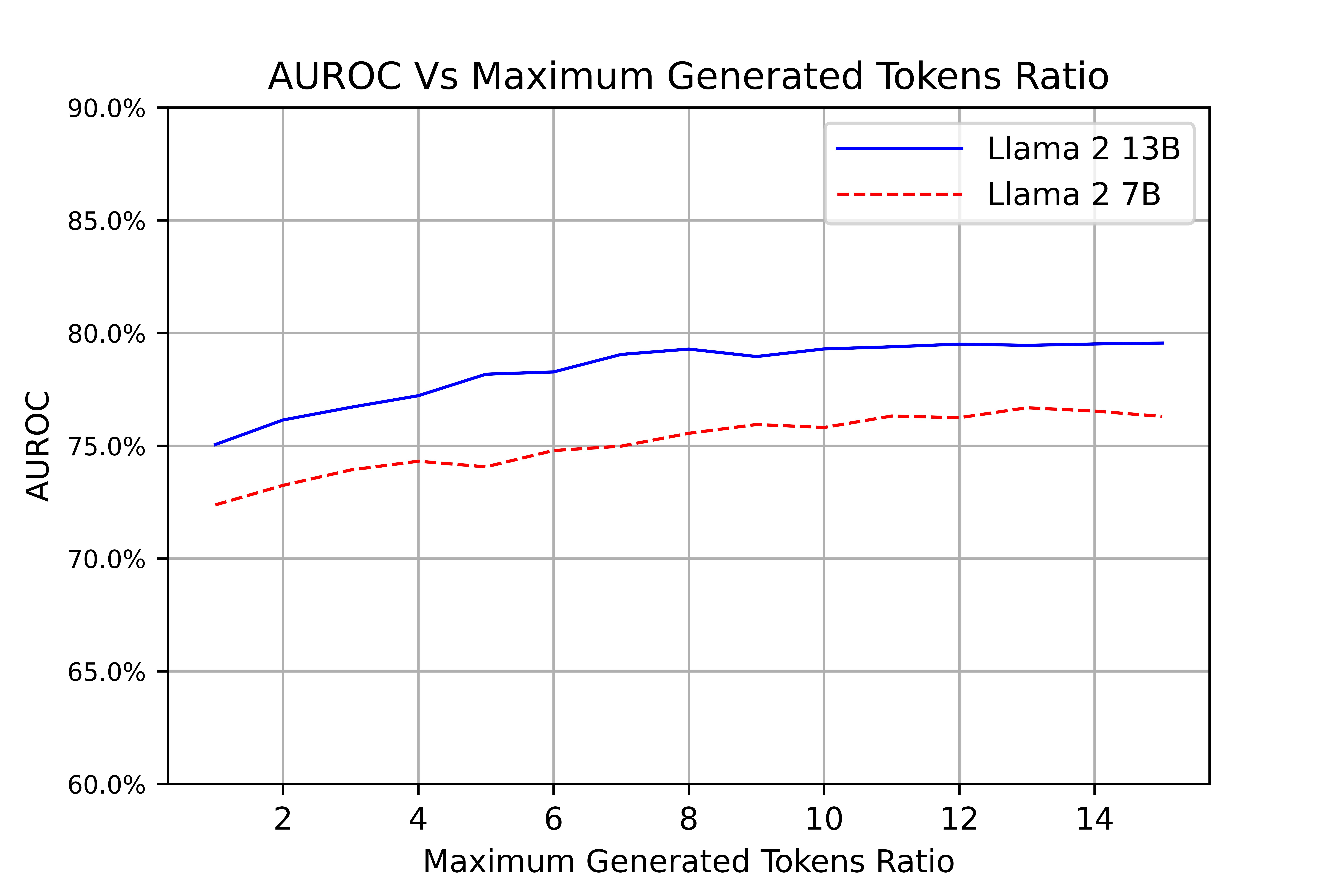}
\caption{Diversion decoding 
AUROC vs maximum generated tokens ratio}
\label{fig:MGTR}
\end{figure}

\subsection{Ablation Studies}

In our experiments, we utilized a training dataset of 1,300 samples and a testing dataset of 3,600 samples, resulting in a training data ratio of approximately 26\%. Despite the limited amount of training data, diversion decoding demonstrated robust performance, achieving AUROC scores of approximately 74\% and 78\% for the Llama 7B and 13B models, respectively. We conducted additional experiments with varying training data ratios to investigate further the extent of improvement that diversion decoding can achieve. Figure~\ref{fig:train_ratio} illustrates the relationship between the AUROC score of diversion decoding and the training data ratio in our dataset. As expected, increasing the amount of training data led to a significant performance boost. Notably, with a 90\% training data ratio, the AUROC score reached approximately 82\% for Llama 7B and around 78\% for Llama 13B.

In our complexity analysis, we report the median expansion ratio of diversion decoding. In practice, LLMs may repeatedly generate the same response upon rejection, leading to a substantial increase in the number of generated tokens. To mitigate this, diversion decoding imposes a constraint on the number of generated tokens beyond the initial response. We represent this number as a ratio to the number of tokens in the initial response and denote it as a maximum generated token ratio (MGTR). More formally: \[
\text{MGTR} = \frac{T_g}{T_i}
\]

Where:
\begin{itemize}
  \item \( T_g \) is the number of tokens generated beyond the initial response.
  \item \( T_i \) is the number of tokens in the initial response.
\end{itemize}

For users concerned with computational complexity, a trade-off exists between performance and token generation: reducing the maximum number of generated tokens can lower complexity while maintaining competitive results. To examine this relationship, we analyzed the impact of MGTR on AUROC scores using a training set ratio of 50\%. The results, presented in Figure \ref{fig:MGTR}, indicate a significant increase in AUROC when MGTR is below 9. Beyond this threshold, the AUROC score plateaus. Notably, even with a relatively low MGTR of 5, diversion decoding achieves competitive AUROC scores of approximately 74\% and 78\% for Llama 2 models with 7B and 13B parameters, respectively.

\section{Related Work}

The problem of hallucination detection and quantification has inspired several approaches. Maynez et al. \cite{maynez2020faithfulness} conducted a large-scale human evaluation on abstractive summarization models, revealing that even fluent summaries often contain hallucinations, and argued for entailment-based metrics over ROUGE to better capture faithfulness. Some studies tackle the problem by fact-checking the output against a verified knowledge base \cite{guo-etal-2022-survey}. However, the size and extensiveness required of the knowledge base can be prohibitive, as relevant evidence must be present in the knowledge base before it can be used to challenge the large language model’s output. Instead of querying an external database, researchers have also explored the possibility of getting the LLM to self-evaluate its confidence regarding the output \cite{kadavath2022language,lin2022teaching}. Other proposed approaches take advantage of the model’s token probabilities or hidden representations to identify sequences in the output that the model is less confident about \cite{azaria-mitchell-2023-internal,fu-etal-2024-gptscore}. Chen et al.\cite{chen2024inside} proposed INSIDE, which leverages the internal states of LLMs to compute an \emph{EigenScore} measuring semantic dispersion across generations and uses feature clipping to reduce overconfident hallucinations. Researchers have also introduced RAGTruth, a specialized corpus for detecting hallucinations in fine-tuned LLM applications that utilize the retrieval-augmented generation (RAG) technique \cite{niu2024ragtruth}.

SelfCheckGPT~\cite{manakul-etal-2023-selfcheckgpt} proposed comparing the LLM output against stochastically generated responses to check for divergence. It leverages the same LLM behavior that we take advantage of, wherein stochastically generated responses containing hallucinated facts are more likely to differ from each other. Uncertainty estimation methods have also demonstrated promising potential in identifying LLM hallucinations across multiple NLP tasks from different domains \cite{huang2023look}. Moreover, researchers introduced semantic uncertainty as a method for quantifying uncertainty in natural language, with the goal of extending beyond token-level probability to consider semantic equivalence among generated responses~\cite{farquhar2024old,Farquhar2024}. Assessing semantic uncertainty requires the following steps: (1) sampling M sequences given context from the predictive distribution of an LLM, (2) utilizing a bidirectional entailment algorithm to categorize semantically equivalent sequences together, and (3) calculating the semantic entropy after summing the probabilities of the sequences that were found to be semantically equivalent from the previous step.



\begin{table}[t]
    \centering
    \renewcommand{\arraystretch}{1.2} 
    \caption{Comparison of different hallucination detection approaches. 
    \textit{"Num. Responses"}: number of responses each approach generates per query; 
    \textit{"Deterministic?"}: whether the approach consistently produces the same output given the same input; 
    \textit{"Retrieval Free?"}: whether the approach functions without reliance on external knowledge bases or RAG techniques; 
    \textit{"Finetuning Free?"}: whether it does not require model finetuning.}
    \begin{tabular}{|>{\centering\arraybackslash\footnotesize}m{25mm}|
                    >{\centering\arraybackslash\footnotesize}m{20mm}|
                    >{\centering\arraybackslash\footnotesize}m{25mm}|
                    >{\centering\arraybackslash\footnotesize}m{20mm}|
                    >{\centering\arraybackslash\footnotesize}m{20mm}|}
        \hline
        \textbf{Approach} & \textbf{Num. Responses} & \textbf{Deterministic?} & \textbf{Retrieval Free?} & \textbf{Finetuning Free?} \\
        \hline
        Semantic Uncertainty \cite{farquhar2024old,Farquhar2024}  & $\sim$10  & \xmark  & \cmark  & \cmark  \\
        \hline
        SelfCheckGPT \cite{manakul-etal-2023-selfcheckgpt}         & $\sim$10  & \xmark  & \cmark  & \cmark  \\
        \hline
        RAGTruth \cite{niu2024ragtruth}             & 1         & \cmark  & \xmark  & \xmark  \\
        \hline
        Diversion Decoding   & $\sim$2   & \cmark  & \cmark  & \cmark  \\
        \hline
    \end{tabular}
    \label{tab:comparison}
\end{table}


Existing approaches exhibit two primary limitations. First, they are computationally expensive. For example, methods such as SelfCheckGPT and semantic entropy require generating multiple response samples (typically around ten) to achieve a reliable output. Given the high computational cost of running large language models (LLMs), these methods demand approximately ten times the resources needed for generating a single response, substantially increasing the cost of assessing uncertainty in LLM outputs. In contrast, our approach is significantly more efficient, requiring only two responses: one generated greedily and another obtained through a slightly more computationally intensive diversion strategy.
Second, existing methods rely on probabilistic techniques, leading to variability in their results across different executions due to stochastic sampling from the response space. This inherent randomness reduces their reliability and practical applicability. In contrast, our approach is deterministic, ensuring consistent outputs without randomness. This determinism enhances interpretability, facilitates debugging, and improves usability, making our method more robust and practical for real-world applications. Table~\ref{tab:comparison} presents a comparison between diversion decoding and various hallucination detection approaches. While diversion decoding requires only two responses, generating the second response is computationally more expensive, as analyzed in the evaluation section.



\section{Conclusion, Limitations and Future Work}
Despite the increasing adoption of LLMs in various applications, there remains a significant security risk of hallucinated LLM outputs going undetected. In this paper, we introduced diversion decoding as a method of developing a proxy measure for the LLMs' uncertainty and subsequently detecting hallucination in the output of LLMs. The approach consists of three main components: a diversion decoder, which forces the LLM to generate a semantically different answer to the greedy response; a semantic similarity module, which prevents the second answer from being similar to the greedy response; and an uncertainty assessment module, which is a trained machine learning model that develops a proxy for the uncertainty of the model. Our experimental results demonstrate that the proposed approach can achieve increased accuracy in the detection of hallucinations compared to other baselines.

Our approach focuses on assessing LLM hallucinations in the context of factual questions with concise answers. While this is a crucial aspect of evaluating language model reliability, LLMs have a wide range of use cases that extend beyond this specific task, including reading comprehension and problem-solving. In future work, we plan to improve diversion decoding techniques to handle a broader range of question formats beyond factual queries.
Moreover, our current approach primarily focuses on open-source models due to their transparency and accessibility. However, with additional components, diversion decoding can also be adapted to closed-source LLMs by leveraging similar features. 
Finally, our approach employs traditional machine learning models, specifically gradient boosting, to detect hallucinations. 
We plan to experiment with more deep learning approaches, such as neural networks, to improve detection accuracy in future work.

\noindent\textbf{Disclaimer} This paper identifies certain equipment, instruments, software, or materials to adequately describe the experimental procedure. Such identification is not intended to imply recommendation or endorsement of any product or
service by NIST, nor is it intended to imply that the materials or equipment identified
are necessarily the best available for the purpose. The use of Llama in our evaluation experiments is primarily because it is an open source model. We do not imply its recommendation or endorsement.

\noindent\textbf{ACKNOWLEDGMENT}
 The research reported herein was supported in part by NIST grant number 60NANB24D143. Any opinions, findings, conclusions, and recommendations expressed in this material are those of the author(s) and do not necessarily reflect the views of NIST.

%
%
%
\clearpage

\bibliographystyle{splncs04}
\bibliography{references}
%




\end{document}